\documentclass[letterpaper]{article} 
\usepackage{aaai2026}
\usepackage{times}  
\usepackage{helvet}  
\usepackage{courier}  
\usepackage[hyphens]{url}  
\usepackage{graphicx} 
\usepackage{natbib}  
\usepackage{caption} 
\usepackage{algorithm}
\usepackage{algorithmicx}

\usepackage{newfloat}
\usepackage{listings}
\DeclareCaptionStyle{ruled}{labelfont=normalfont,labelsep=colon,strut=off} 
\floatstyle{ruled}
\newfloat{listing}{tb}{lst}{}
\floatname{listing}{Listing}
\usepackage{booktabs}       
\usepackage{multirow}       
\usepackage{makecell}       
\usepackage[table]{xcolor}  
\usepackage{subcaption}     
\usepackage{algpseudocode}  
\usepackage{amsmath}        
\graphicspath{
    {figures/}   
    {./}         
}

\newcommand{\perf}[2]{\shortstack{#1 \\ \small (#2)}}
\newcommand{\perfbest}[2]{\textbf{\underline{\perf{#1}{#2}}}} 
\newcommand{\perfsecond}[2]{\shortstack{#1 \\ \small (#2)}}   
\newcommand{\perforiginal}[2]{\textbf{\perf{#1}{#2}}}      
\newcommand{\peravgbest}[1]{\textbf{\underline{\small #1}}}   
\newcommand{\peravg}[1]{\small #1}                      
\definecolor{ourscell}{gray}{0.9}  
\newcommand{\bestperf}[1]{\textbf{\underline{#1}}}
\nocopyright 

\title{ViTCoP: Accelerating Large Vision-Language Models via Visual and Textual Semantic Collaborative Pruning}
\author{
    Wen Luo,
    Peng Chen,
    Xiaotao Huang\thanks{Corresponding Author.},
    LiQun Huang\footnotemark[1]
}
\affiliations{
    School of Software Engineering, Huazhong University of Science and Technology, WuHan, China\\
    \{wen\_chaser, hustchenpeng\}@hust.edu.cn, \{huangxiaotao, huangliqun\}@mail.hust.edu.cn
}

\usepackage{bibentry}

\begin{document}

\maketitle

\begin{abstract}
Large Vision-Language Models (LVLMs) incur high computational costs due to significant redundancy in their visual tokens. To effectively reduce this cost, researchers have proposed various visual token pruning methods. However, existing methods are generally limited, either losing critical visual information prematurely due to pruning in the vision encoder, or leading to information redundancy among the selected tokens due to pruning in the Large Language Models (LLMs). To address these challenges, we propose a Visual and Textual Semantic Collaborative Pruning framework (ViTCoP) that combines redundancy filtering in the vision encoder with step-wise co-pruning within the LLM based on its hierarchical characteristics, to efficiently preserve critical and informationally diverse visual tokens. Meanwhile, to ensure compatibility with acceleration techniques like FlashAttention, we introduce the L2 norm of K-vectors as the token saliency metric in the LLM. Extensive experiments on various Large Vision-Language Models demonstrate that ViTCoP not only achieves state-of-the-art performance surpassing existing methods on both image and video understanding tasks, but also significantly reduces model inference latency and GPU memory consumption. Notably, its performance advantage over other methods becomes even more pronounced under extreme pruning rates.
\end{abstract}

\begin{links}
\link{Code}{https://github.com/chaser682/ViTCoP}
\end{links}

\section{Introduction}
\label{sec:introduction}

\begin{figure}[ht]
    \centering
    \includegraphics[width=1.0\linewidth]{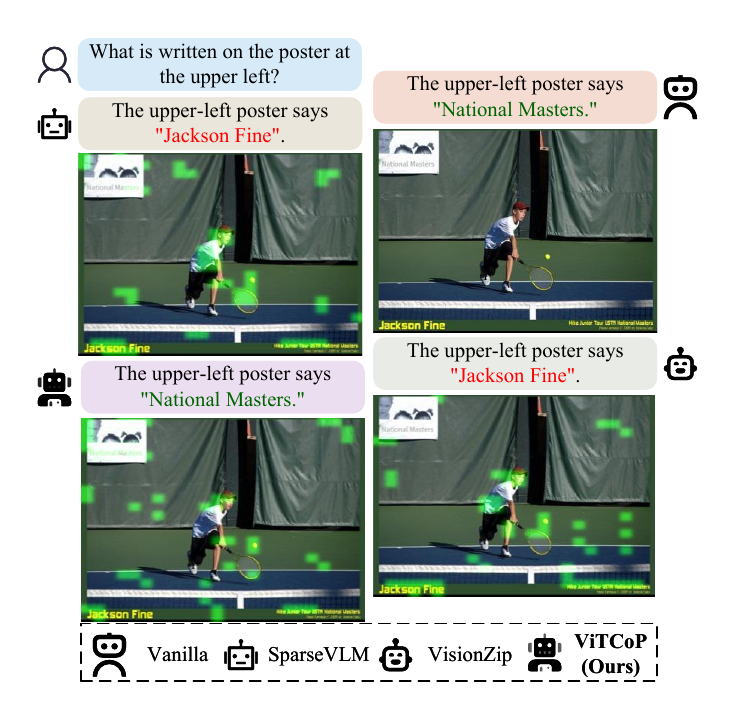}
    \caption{
        Visual question answering results of LLaVA-1.5-7B with different pruning methods.
    }
    \label{fig:method_comparison}
\end{figure}

The monumental success of Large Language Models (LLMs) in the domain of language understanding~\cite{openai2024gpt4technicalreport,vicuna2023,Touvron2023LLaMAOA,qwen2025qwen25technicalreport} has catalyzed the proliferation and remarkable advancement of Large Vision-Language Models (LVLMs).
LVLMs~\cite{lin2024videollavalearningunitedvisual,liu2023visualinstructiontuning,liu2024llavanext,zhang2024llavanextvideo} operate by encoding visual information from images and videos into a vast number of visual tokens. Through a lightweight modality-alignment module~\cite{liu2023visualinstructiontuning,bai2025qwen25vltechnicalreport,li2023blip2bootstrappinglanguageimagepretraining}, these visual tokens are concatenated with text tokens and subsequently fed into an LLM for instruction fine-tuning~\cite{liu2023visualinstructiontuning}.
This paradigm has endowed LVLMs with powerful multimodal perception and reasoning capabilities across a spectrum of tasks, including image comprehension and video question-answering.

However, despite their exceptional performance, the substantial computational cost of LVLMs presents a critical bottleneck.
The inherent density of visual information, particularly in high-resolution images or long videos, results in the generation of thousands, or even tens of thousands, of visual tokens~\cite{zhang2024internlmxcomposer25versatilelargevision,chen2024sharegpt4videoimprovingvideounderstanding,maaz2024videochatgptdetailedvideounderstanding}.
Given that the computational complexity of the Transformer architecture scales quadratically with the input sequence length, this deluge of visual tokens leads to prohibitive inference latency and GPU memory consumption.
This overhead severely constrains the efficient deployment and application of LVLMs in resource-constrained environments such as autonomous driving, robotics, and edge computing~\cite{kim2024openvlaopensourcevisionlanguageactionmodel,liu2024robomambaefficientvisionlanguageactionmodel,qu2025mobileedgeintelligencelarge,yang2024unifiedlanguagedrivenzeroshotdomain,yao2024minicpmvgpt4vlevelmllm}.

Existing research indicates that a high degree of information redundancy exists among the visual tokens in LVLMs~\cite{chen2024imageworth12tokens,shang2024llavaprumergeadaptivetokenreduction,xing2025pyramiddropacceleratinglargevisionlanguage,zhang2025sparsevlmvisualtokensparsification,yang2024visionziplongerbetternecessary}. To address this challenge, visual token pruning has emerged as a promising technical direction, with current work broadly categorized into two paradigms. The first is text-agnostic pruning, which operates solely on visual information without considering the specific text instruction. For instance, VisionZip~\cite{yang2024visionziplongerbetternecessary} identifies dominant tokens via attention scores and employs a token fusion strategy to extract contextually rich representations. The fundamental limitation of such methods, however, is their disregard for guidance from the language instruction. As illustrated in Figure~\ref{fig:method_comparison}, when asked "What is written on the poster at the upper left?", a text-agnostic method like VisionZip retains many visually salient tokens from the player and court, but may fail to focus on the specific poster, leading to an incorrect answer.Since user queries often pertain to specific regions, this text-agnostic strategy may preserve task-irrelevant visual information, degrading model performance.The second category, text-guided pruning, leverages the textual instruction to direct the process. Methods like FastV~\cite{chen2024imageworth12tokens} and PyramidDrop~\cite{xing2025pyramiddropacceleratinglargevisionlanguage} use text-attention scores to identify and discard unimportant tokens, but this may leads to high redundancy among the selected tokens.Similarly, SparseVLM~\cite{zhang2025sparsevlmvisualtokensparsification} employs visually-relevant text tokens as raters to filter for important visual tokens. However, it also has its limitations. When text instructions are broad or focus on similar concepts, the visual tokens selected under this guidance may exhibit significant content overlap, leading to high information redundancy and insufficient diversity.Consequently, existing methods face a significant challenge: purely visual pruning risks losing critical details, while purely text-guided pruning in the LLM tends to yield high informational redundancy.

To resolve these challenges, we propose ViTCoP, a Visual-Text Collaborative Pruning framework.
Our core insight is that an optimal pruning strategy must synergistically leverage semantic information from different modalities at distinct stages of the LVLM's processing pipeline.
To this end, ViTCoP employs an innovative three-stage strategy.
First, within the vision encoder, we perform a coarse-grained, visually-guided pruning to remove patently redundant tokens from backgrounds or repetitive textures. Second, in the shallow layers of the LLM, where the model performs initial global cross-modal understanding~\cite{neo2025interpretingvisualinformationprocessing,zhang2024redundancyrelevanceinformationflow}, we employ a vision-text synergistic pruning to ensure the retained tokens are both highly relevant to the query and semantically diverse.
Finally, in the deep layers of the LLM, as the model's understanding of the instruction becomes progressively more focused~\cite{parekh2024conceptbasedexplainabilityframeworklarge,chen2024imageworth12tokens,xing2025pyramiddropacceleratinglargevisionlanguage}, we transition to a text-guided, fine-grained pruning to further refine the selection down to the core visual evidence most directly pertinent to the final answer.

\begin{figure}[ht]
    \centering
    \begin{subfigure}[b]{0.49\linewidth}
        \includegraphics[width=\linewidth]{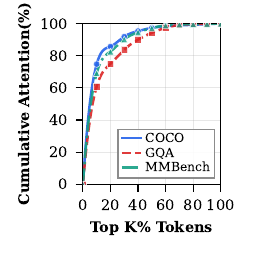}
        \caption{Attention Cumulative Distribution}
        \label{fig:attn_cdf}
    \end{subfigure}
    \hfill
    \begin{subfigure}[b]{0.49\linewidth}
        \includegraphics[width=\linewidth]{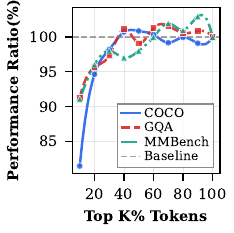}
        \caption{Performance vs. Top K\% Token}
        \label{fig:perf_retention}
    \end{subfigure}
    \caption{Analysis of initial visual token redundancy. (a) A small fraction of tokens captures a majority of the attention score. (b) Model performance shows minimal degradation even when a large portion of tokens is pruned.}
    \label{fig:initial_redundancy}
\end{figure}

Through this hierarchical and progressive strategy, ViTCoP adeptly balances the preservation of critical information with the promotion of token diversity.
Furthermore, to ensure compatibility with modern acceleration techniques such as FlashAttention~\cite{dao2022flashattentionfastmemoryefficientexact,dao2023flashattention2fasterattentionbetter}, we innovatively introduce the L2 norm of key vectors as a lightweight yet effective saliency metric for token selection in LVLMs.
Extensive experiments on multiple mainstream LVLMs demonstrate that ViTCoP not only achieves state-of-the-art performance on image and video understanding benchmarks but also significantly reduces inference latency and GPU memory footprint. 


\section{Insights}
\label{sec:insights}

\subsection{Initial Redundancy of Visual Tokens}

Our study reveals significant initial redundancy in visual tokens generated by the Vision Transformer. On the LLaVA-1.5-7B model \cite{liu2023visualinstructiontuning}, we found that the top 10\% of tokens with the highest attention scores contribute over 60\% of the total attention weight (Figure~\ref{fig:initial_redundancy}a). More importantly, retaining just the top 20\% of tokens is sufficient to maintain approximately 95\% of the model's performance across various image-language understanding benchmarks (Figure~\ref{fig:initial_redundancy}b). This confirms that a small subset of visual tokens can represent the vast majority of an image's information.

\textbf{Key Insight 1:} A large number of visual tokens can be pruned before entering the LLM with minimal impact on model performance.

\subsection{K-Vector L2 Norm: An Efficient Proxy for Token Saliency}

\begin{figure}[ht]
    \centering
    \begin{subfigure}[b]{0.49\linewidth}
        \includegraphics[width=\linewidth, height=3.5cm]{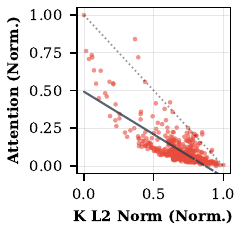}
        \caption{Correlation Analysis}
        \label{fig:l2_attn_corr}
    \end{subfigure}
    \hfill
    \begin{subfigure}[b]{0.49\linewidth}
        \includegraphics[width=\linewidth, height=3.5cm]{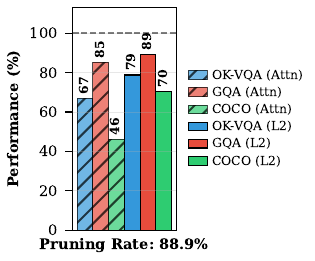}
        \caption{Performance Comparison}
        \label{fig:l2_attn_perf}
    \end{subfigure}
    \caption{Validation of K-vector L2 norm as a saliency proxy. (a) A strong negative correlation exists between L2 norm and attention. (b) L2 norm-based pruning is competitive with, or superior to, attention-based methods.}
    \label{fig:l2_norm_proxy}
\end{figure}

Pruning based on attention scores, as used in methods like FastV~\cite{chen2024imageworth12tokens}, is effective but often incompatible with modern computational optimizations like FlashAttention~\cite{dao2022flashattentionfastmemoryefficientexact,dao2023flashattention2fasterattentionbetter}. Inspired by recent work~\cite{devoto2024simpleeffective}, we investigate the L2 norm of Key (K) vectors as a lightweight proxy. Our analysis reveals a strong negative correlation between the K-vector L2 norm and attention scores (Figure~\ref{fig:l2_norm_proxy}a). Furthermore, comparative experiments show that pruning based on the smallest L2 norm achieves performance that is competitive with, and at times superior to, attention-based pruning across multiple benchmarks (Figure~\ref{fig:l2_norm_proxy}b).

\textbf{Key Insight 2:} In LVLMs, the K-vector L2 norm is a lightweight and effective proxy for token saliency within the LLM, where a smaller norm corresponds to higher importance.

\subsection{Evolving Importance of Visual Tokens in LLM}

\begin{figure}[ht]
    \centering
    \includegraphics[width=1.0\linewidth]{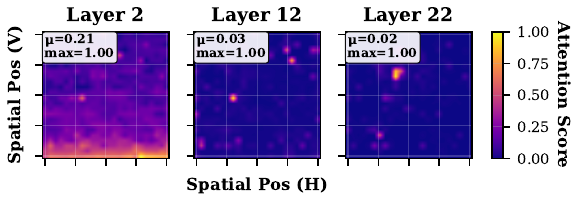}
    \caption{Heatmap of visual token attention scores across LLM layers. }
    \label{fig:attention_heatmap}
\end{figure}

The importance of visual tokens is not static but evolves as they propagate through the LLM layers. By analyzing the distribution of attention scores across different layers (Figure~\ref{fig:attention_heatmap}), we observe a clear functional shift: the LLM transitions from aggregating diverse, global information in the shallow layers to focusing on key local details in the deep layers.

\textbf{Key Insight 3:} The LLM aggregates global visual information in shallow layers and focuses on absorbing key local visual information in deep layers.

\section{Method}
\label{sec:method}

\begin{figure*}[ht]
    \centering
    \includegraphics[width=1.0\textwidth, height=8cm]{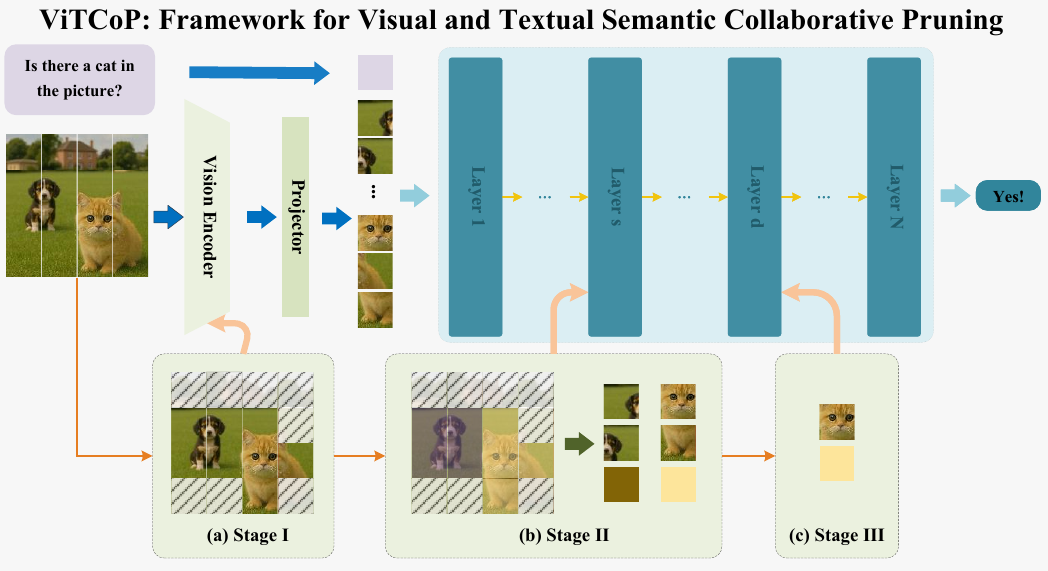}
    \caption{The ViTCoP framework's three-stage process: (a) Coarse pruning in the Vision Encoder via `[CLS]` attention, (b) Collaborative pruning in shallow LLM layers using VIC clustering and K-norm merging, and (c) Aggressive text-saliency pruning in deep LLM layers.}
    \label{fig:vitcop_framework}
\end{figure*}

In this paper, we propose ViTCoP, a dynamic token pruning framework based on Visual-Textual Semantic Collaborative Pruning. The core strategy of ViTCoP is to synergistically leverage visual-textual semantic information to perform a multi-stage, differentiated pruning adapted to the different phases of a LVLM.
As illustrated in Figure~\ref{fig:vitcop_framework}, ViTCoP consists of three stages: (I) Coarse-grained pruning guided by visual saliency in the vision encoder; (II) Collaborative visual-textual semantic-guided pruning in the shallow layers of the LLM to acquire tokens that are both semantically diverse and text-relevant; and (III) Fine-grained pruning guided by textual saliency in the deep layers of the LLM. Through this synergistic visual-textual pruning strategy, ViTCoP strikes a balance between preserving critical and diverse information.

\subsection{Stage I: Visual Saliency-Guided Pruning in the Vision Encoder}
As discussed in Section 2.1, a significant number of redundant tokens already exist in the vision encoder. Therefore, this initial stage aims to eliminate highly redundant tokens, such as those from information-sparse backgrounds or repetitive textures, to provide a high-quality input for the subsequent fine-grained pruning within the LLM. Specifically, for the visual tokens entering the LVLM's projection layer (including the \texttt{[CLS]} token in CLIP~\cite{radford2021learningtransferablevisualmodels}), we define the saliency score of the $i$-th visual token based on the attention it receives from the \texttt{[CLS]} token:
\begin{equation}
\label{eq:saliency_score}
S_i = \sum_{h=1}^{H} \mathbf{A}^{(h)}_{0, i},
\end{equation}
where $H$ is the number of attention heads, and $\mathbf{A}^{(h)}_{0, i}$ represents the attention score from the \texttt{[CLS]} token (at index $0$) to the $i$-th visual token in the $h$-th attention head. By ranking the visual tokens based on their saliency scores $S_i$ and selecting the top-ranking ones, this stage preserves high-saliency tokens rich in information, removing useless redundancy for the subsequent pruning stages in the LLM.

\subsection{Stage II: Visual-Textual Collaborative Pruning in Shallow LLM Layers}
As established in Section 2.3, the LLM needs to perform a preliminary global understanding by integrating both visual and textual information in its shallow layers. Therefore, we employ a collaborative visual-textual semantic-guided pruning strategy to ensure that the retained tokens are not only semantically diverse but also highly relevant to the text.

\subsubsection{Visual Semantic Guidance: VIC Algorithm}
For visual semantic guidance, we introduce the Visual Information Clustering (VIC) algorithm, designed to preserve the diversity of visual semantic information. Specifically, the inputs to VIC are the feature vectors of the high-saliency tokens retained from Stage I and their corresponding position vectors in the original image. The output of our algorithm depends on three parameters: a cutoff distance ($d_c$), a spatial threshold ($\tau$), and a ratio of cluster centers. We calculate feature and spatial distances, and the local density $\rho_i$ for each token $i$ is computed as:
\begin{equation}
\label{eq:local_density}
\rho_i = \sum_{j \neq i} \exp\left(-\left(\frac{d_{ij}}{d_c}\right)^2\right),
\end{equation}
where $d_{ij}$ denotes the feature distance between tokens $i$ and $j$.

For each token $i$, we find the minimum feature distance $\delta_i$ to another token $j$ that has a higher density ($\rho_j > \rho_i$) and is within the spatial distance threshold $\tau$:
\begin{equation}
\label{eq:min_distance}
\delta_i = \min_{\substack{j: \rho_j > \rho_i \\ d_{\text{spatial}}(i,j) \leq \tau}} d_{ij},
\end{equation}
where $d_{\text{spatial}}(i,j)$ represents the spatial distance between tokens $i$ and $j$.

We then calculate an importance score $\gamma_i = \rho_i \cdot \delta_i$ for each token, where tokens with the highest importance scores are designated as cluster centers. Subsequently, each non-center token is assigned to the cluster of its nearest center. Our algorithm ensures that each token is clustered into a semantically coherent group, thereby satisfying the subsequent need to retain semantically diverse tokens. 

\subsubsection{Textual Semantic Guidance}
As noted in Section 2.2, the L2 norm of the Key (K) vectors exhibits a strong negative correlation with attention scores. That is, visual tokens more relevant to the text tend to have smaller K-vector L2 norms. Therefore, we use the L2 norm of the K vectors from the LLM's attention module as a token saliency metric. The L2 norm of a token's K vector is calculated as:
\begin{equation}
\label{eq:k_vector_norm}
\|\mathbf{K}_i\|_2 = \sqrt{\sum_{h=1}^{H} \|\mathbf{K}_i^{(h)}\|_2^2},
\end{equation}
where $H$ is the number of attention heads and $\mathbf{K}_i^{(h)}$ is the K vector of the $i$-th token in the $h$-th head.

\subsubsection{Collaborative Pruning and Merging}
To achieve collaborative pruning guided by both visual and textual semantics, we proceed as follows. Given a set of visual tokens with their cluster labels from the VIC algorithm and their K-vector L2 norms, we first assign a retention quota $q_c$ to each cluster $c$. This quota determines the number of elite tokens to be retained from that cluster and is proportional to the cluster's relative size, ensuring minimal information loss:
\begin{equation}
\label{eq:retention_quota}
q_c = \left\lfloor \frac{|C_c|}{\sum_{k=1}^{N_c} |C_k|} \cdot (B - N_c) \right\rfloor,
\end{equation}
where $B$ is the total budget for elite tokens, $|C_c|$ is the size of cluster $c$, and $N_c$ is the total number of clusters. For the selection of elite tokens within each cluster, we select the top $q_c$ tokens with the smallest K-vector L2 norms, as a smaller norm indicates higher relevance to the text. Finally, the remaining tokens within each cluster are merged into a single representative token by averaging their feature vectors:
\begin{equation}
\label{eq:token_merging}
\mathbf{t}_c^{\text{merged}} = \frac{1}{|C_c^{\text{remaining}}|} \sum_{i \in C_c^{\text{remaining}}} \mathbf{t}_i,
\end{equation}
where $C_c^{\text{remaining}}$ denotes the set of remaining tokens in cluster $c$ after elite selection, and $\mathbf{t}_i$ represents the feature vector of token $i$. This collaborative approach ensures that both fine-grained details and generalized context are preserved. 

\subsection{Stage III: Textual Saliency-Guided Pruning in Deep LLM Layers}

\begin{table*}[ht]
 \centering
 \small
 \renewcommand{\arraystretch}{0.7}
 \setlength{\tabcolsep}{0.8pt}

 \begin{tabular}{lccccccccccc c}
 \toprule
 \textbf{Method} & \textbf{COCO} & \textbf{Flickr} & \textbf{GQA} & \textbf{MMB} & \textbf{MME} & \textbf{NoCaps} & \textbf{OK-VQA} & \textbf{POPE} & \textbf{QBench} & \textbf{SQA} & \textbf{VQA-v2} & \textbf{Avg (\%)} \\
 \midrule
 \textbf{Vanilla} & \perforiginal{1.102}{100.0\%} & \perforiginal{0.750}{100.0\%} & \perforiginal{0.619}{100.0\%} & \perforiginal{64.08}{100.0\%} & \perforiginal{1862}{100.0\%} & \perforiginal{1.055}{100.0\%} & \perforiginal{0.534}{100.0\%} & \perforiginal{0.858}{100.0\%} & \perforiginal{0.585}{100.0\%} & \perforiginal{0.695}{100.0\%} & \perforiginal{0.716}{100.0\%} & \textbf{100.0\%} \\
 \midrule
 \multicolumn{13}{c}{\textbf{Retain 192 Tokens {(↓ 66.7\%)}}} \\
 \midrule
 FastV & \perf{1.082}{98.1\%} & \perfbest{0.741}{98.7\%} & \perf{0.527}{85.1\%} & \perf{60.57}{94.5\%} & \perf{1612}{86.6\%} & \perfbest{1.033}{97.9\%} & \perf{0.512}{95.9\%} & \perf{0.646}{75.3\%} & \perfsecond{0.581}{99.3\%} & \perf{0.672}{96.7\%} & \perf{0.663}{92.6\%} & \peravg{92.8\%} \\
 PyramidDrop & \perfbest{1.091}{99.0\%} & \perf{0.734}{97.9\%} & \perf{0.574}{92.7\%} & \perfsecond{63.75}{99.5\%} & \perfsecond{1797}{96.5\%} & \perfsecond{1.023}{97.0\%} & \perf{0.508}{95.1\%} & \perf{0.810}{94.4\%} & \perfbest{0.581}{99.3\%} & \perfbest{0.692}{99.6\%} & \perf{0.678}{94.7\%} & \peravg{96.9\%} \\
 SparseVLM & \perfsecond{1.087}{98.6\%} & \perf{0.720}{95.9\%} & \perf{0.576}{93.0\%} & \perf{62.92}{98.2\%} & \perf{1721}{92.4\%} & \perf{1.010}{95.7\%} & \perf{0.520}{97.4\%} & \perf{0.837}{97.5\%} & \perf{0.575}{98.3\%} & \perfsecond{0.692}{99.6\%} & \perfbest{0.706}{98.6\%} & \peravg{96.8\%} \\
 VisionZip & \perf{1.070}{97.0\%} & \perfsecond{0.737}{98.3\%} & \perfsecond{0.593}{95.8\%} & \perf{63.66}{99.3\%} & \perf{1782}{95.7\%} & \perf{1.023}{97.0\%} & \perfsecond{0.525}{98.3\%} & \perfsecond{0.853}{99.4\%} & \perf{0.575}{98.3\%} & \perf{0.689}{99.1\%} & \perf{0.686}{95.8\%} & \peravg{97.6\%} \\
 \cmidrule(lr){2-13}
 \rowcolor{ourscell}
 \textbf{ViTCoP (Ours)} & \perf{1.078}{97.8\%} & \perf{0.735}{98.0\%} & \perfbest{0.600}{96.9\%} & \perfbest{64.26}{100.3\%} & \perfbest{1816}{97.5\%} & \perf{1.019}{96.6\%} & \perfbest{0.536}{100.4\%} & \perfbest{0.855}{99.6\%} & \perf{0.579}{99.0\%} & \perf{0.684}{98.4\%} & \perfsecond{0.705}{98.5\%} & \peravgbest{98.5\%} \\
 \midrule[1pt]
 \multicolumn{13}{c}{\textbf{Retain 128 Tokens {(↓ 77.8\%)}}} \\
 \midrule
 FastV & \perfsecond{1.044}{94.7\%} & \perfsecond{0.719}{95.8\%} & \perf{0.496}{80.1\%} & \perf{57.29}{89.4\%} & \perf{1490}{80.0\%} & \perfsecond{0.995}{94.3\%} & \perf{0.486}{91.0\%} & \perf{0.597}{69.5\%} & \perfsecond{0.579}{99.0\%} & \perf{0.602}{86.6\%} & \perf{0.632}{88.3\%} & \peravg{88.9\%} \\
 PyramidDrop & \perf{1.039}{94.2\%} & \perf{0.692}{92.2\%} & \perf{0.572}{92.4\%} & \perf{59.89}{93.5\%} & \perfsecond{1761}{94.6\%} & \perf{0.969}{91.8\%} & \perf{0.491}{91.9\%} & \perf{0.738}{86.0\%} & \perfbest{0.581}{99.3\%} & \perf{0.684}{98.4\%} & \perf{0.650}{90.8\%} & \peravg{93.2\%} \\
 SparseVLM & \perf{0.940}{85.3\%} & \perf{0.583}{77.7\%} & \perf{0.561}{90.6\%} & \perf{60.71}{94.7\%} & \perf{1696}{91.1\%} & \perf{0.823}{78.0\%} & \perfsecond{0.509}{95.3\%} & \perf{0.805}{93.8\%} & \perf{0.572}{97.8\%} & \perf{0.672}{96.7\%} & \perfbest{0.684}{95.5\%} & \peravg{90.6\%} \\
 VisionZip & \perf{1.037}{94.1\%} & \perf{0.713}{95.1\%} & \perfsecond{0.576}{93.0\%} & \perfsecond{62.37}{97.3\%} & \perfsecond{1761}{94.6\%} & \perf{0.989}{93.7\%} & \perf{0.507}{95.0\%} & \perfsecond{0.833}{97.1\%} & \perf{0.570}{97.4\%} & \perfbest{0.689}{99.1\%} & \perf{0.665}{92.9\%} & \peravg{95.4\%} \\
 \cmidrule(lr){2-13}
 \rowcolor{ourscell}
 \textbf{ViTCoP (Ours)} & \perfbest{1.064}{96.5\%} & \perfbest{0.724}{96.5\%} & \perfbest{0.592}{95.6\%} & \perfbest{63.83}{99.6\%} & \perfbest{1785}{95.9\%} & \perfbest{1.008}{95.5\%} & \perfbest{0.531}{99.4\%} & \perfbest{0.846}{98.6\%} & \perf{0.577}{98.6\%} & \perfsecond{0.684}{98.4\%} & \perfsecond{0.682}{95.2\%} & \peravgbest{97.3\%} \\
 \midrule[1pt]
 \multicolumn{13}{c}{\textbf{Retain 64 Tokens {(↓ 88.9\%)}}} \\
 \midrule
 FastV & \perf{0.815}{73.9\%} & \perf{0.511}{68.1\%} & \perf{0.462}{74.6\%} & \perf{50.43}{78.7\%} & \perf{1256}{67.5\%} & \perf{0.768}{72.8\%} & \perf{0.370}{69.3\%} & \perf{0.483}{56.3\%} & \perf{0.540}{92.3\%} & \perf{0.512}{73.7\%} & \perf{0.503}{70.2\%} & \peravg{72.5\%} \\
 PyramidDrop & \perf{0.648}{58.8\%} & \perf{0.372}{49.6\%} & \perf{0.475}{76.7\%} & \perf{56.10}{87.5\%} & \perf{1561}{83.8\%} & \perf{0.627}{59.4\%} & \perf{0.395}{74.0\%} & \perf{0.692}{80.6\%} & \perf{0.551}{94.2\%} & \perf{0.608}{87.5\%} & \perf{0.578}{80.7\%} & \peravg{76.6\%} \\
 SparseVLM & \perf{0.731}{66.3\%} & \perf{0.419}{55.9\%} & \perf{0.527}{85.1\%} & \perf{57.90}{90.4\%} & \perf{1505}{80.8\%} & \perf{0.584}{55.4\%} & \perf{0.451}{84.5\%} & \perf{0.758}{88.3\%} & \perfsecond{0.563}{96.2\%} & \perf{0.622}{89.5\%} & \perf{0.615}{85.9\%} & \peravg{80.7\%} \\
 VisionZip & \perfsecond{0.948}{86.0\%} & \perfsecond{0.651}{86.8\%} & \perfsecond{0.551}{89.0\%} & \perfsecond{60.31}{94.1\%} & \perfsecond{1690}{90.8\%} & \perfsecond{0.900}{85.3\%} & \perfsecond{0.478}{89.5\%} & \perfsecond{0.771}{89.9\%} & \perf{0.559}{95.6\%} & \perfbest{0.690}{99.3\%} & \perfsecond{0.631}{88.1\%} & \peravg{90.4\%} \\
 \cmidrule(lr){2-13}
 \rowcolor{ourscell}
 \textbf{ViTCoP (Ours)} & \perfbest{1.032}{93.6\%} & \perfbest{0.696}{92.8\%} & \perfbest{0.574}{92.7\%} & \perfbest{63.06}{98.4\%} & \perfbest{1744}{93.7\%} & \perfbest{0.973}{92.2\%} & \perfbest{0.508}{95.1\%} & \perfbest{0.807}{94.1\%} & \perfbest{0.568}{97.1\%} & \perfsecond{0.688}{99.0\%} & \perfbest{0.663}{92.6\%} & \peravgbest{94.7\%} \\
 \bottomrule
 \end{tabular}
  \caption{Performance on LLaVA-1.5-7B. Each cell shows the score and retention rate (\%). The \textbf{\underline{best}} result in each group is highlighted.}
 \label{tab:pruning_comparison_llava_1_5}
\end{table*}

Once the token sequence propagates to the deep layers of the LLM, the model has progressively absorbed a substantial amount of semantic information from the visual tokens. As per Section 2.3, the LLM in its deep layers focuses on assimilating key local visual information. As the model's understanding of visual information deepens and becomes more focused, a high degree of redundancy emerges among the visual tokens because their core information has been effectively captured. Therefore, we employ a text-saliency-only guided pruning in the deep LLM layers to eliminate a large number of visual tokens that are either irrelevant to the text or whose information has already been aggregated and understood by the model. Specifically, we use the L2 norm of the visual token's K vectors (as defined in Eq.~\ref{eq:k_vector_norm}) to retain the top-ranking salient tokens that contain key local information.

This three-stage, coarse-to-fine filtering significantly enhances ViTCoP's efficiency while maintaining performance.

\section{Experiments}
\label{experiments}

\subsection{Experimental Settings}

\subsubsection{Baselines and Models}
To evaluate the effectiveness of our proposed ViTCoP framework, we compare it against four recent and competitive token pruning baselines: FastV\cite{chen2024imageworth12tokens}, PyramidDrop\cite{xing2025pyramiddropacceleratinglargevisionlanguage}, SparseVLM\cite{zhang2025sparsevlmvisualtokensparsification}, and VisionZip\cite{yang2024visionziplongerbetternecessary}. Our experiments are conducted on a suite of LVLMs to demonstrate its broad applicability. Specifically, we use LLaVA-1.5-7B \cite{liu2023visualinstructiontuning}for image task evaluation, and the more advanced LLaVA-NeXT-7B\cite{liu2024llavanext} and LLaVA-NeXT-Video-7B\cite{zhang2024llavanextvideo} for high-resolution image and video evaluations, respectively.

\begin{table}[ht]
\centering
\small
\renewcommand{\arraystretch}{0.8}
\setlength{\tabcolsep}{0.8pt}

\begin{tabular}{lcccc c}
\toprule
\textbf{Method} & \textbf{COCO} & \textbf{GQA} & \textbf{MMB} & \textbf{POPE} & \textbf{Avg(\%)} \\
\midrule
\textbf{Vanilla} & \perforiginal{1.000}{100.0\%} & \perforiginal{0.643}{100.0\%} & \perforiginal{67.01}{100.0\%} & \perforiginal{0.865}{100.0\%} & \textbf{100.0\%} \\
\midrule
\multicolumn{6}{c}{\textbf{Retain 320 Tokens {(↓ 88.9\%)}}} \\
\midrule
FastV & \perf{0.629}{62.9\%} & \perf{0.533}{82.9\%} & \perf{58.68}{87.6\%} & \perf{0.599}{69.2\%} & \peravg{75.7\%} \\
SparseVLM & \perfsecond{0.839}{83.9\%} & \perf{0.578}{89.9\%} & \perfbest{64.78}{96.7\%} & \perfsecond{0.827}{95.7\%} & \peravg{91.6\%} \\
PyramidDrop & \perf{0.625}{62.5\%} & \perf{0.375}{58.3\%} & \perf{59.36}{88.6\%} & \perf{0.659}{76.2\%} & \peravg{71.4\%} \\
VisionZip & \perf{0.826}{82.6\%} & \perfsecond{0.593}{92.2\%} & \perfsecond{63.83}{95.2\%} & \perf{0.824}{95.3\%} & \peravg{91.4\%} \\
\cmidrule(lr){2-6}
\rowcolor{ourscell}
\textbf{ViTCoP (Ours)} & \perfbest{0.912}{91.2\%} & \perfbest{0.610}{94.9\%} & \perfbest{64.78}{96.7\%} & \perfbest{0.846}{97.8\%} & \peravgbest{95.1\%} \\
\midrule
\multicolumn{6}{c}{\textbf{Retain 160 Tokens {(↓ 94.4\%) \textsuperscript{*}}}} \\
\midrule
VisionZip & \perfsecond{0.697}{69.7\%} & \perfsecond{0.556}{86.5\%} & \perfsecond{60.05}{89.6\%} & \perfsecond{0.757}{87.5\%} & \peravg{83.3\%} \\
\cmidrule(lr){2-6}
\rowcolor{ourscell}
\textbf{ViTCoP (Ours)} & \perfbest{0.844}{84.4\%} & \perfbest{0.584}{90.8\%} & \perfbest{62.89}{93.8\%} & \perfbest{0.816}{94.3\%} & \peravgbest{90.8\%} \\
\bottomrule
\end{tabular}
\caption{Performance comparison on 4 key datasets from LLaVA-NeXT-7B. \textsuperscript{*}At 94.4\% compression, some methods are omitted due to incompatibility.}
\label{tab:pruning_comparison_llava_next}
\end{table}

\subsubsection{Datasets}
Our evaluation covers a wide range of standard benchmarks to ensure a comprehensive assessment of performance across both image and video understanding tasks. For the image-language evaluation, we used 11 diverse datasets: COCO-2017\cite{lin2015microsoft}, Flickr30k\cite{young-etal-2014-image}, GQA\cite{hudson2019gqa}, MMBench\cite{liu2024mmbenchmultimodalmodelallaround}, MME\cite{fu2024mmecomprehensiveevaluationbenchmark}, Nocaps\cite{Agrawal_2019}, OK-VQA\cite{marino2019okvqavisualquestionanswering}, POPE\cite{li2023evaluating}, QBench\cite{wu2024qbenchbenchmarkgeneralpurposefoundation}, ScienceQA\cite{lu2022learn}, and VQA-v2\cite{8100153}. For the video-language evaluation, we utilized 4 representative datasets: EgoSchema\cite{mangalam2023egoschemadiagnosticbenchmarklongform}, MVBench\cite{li2024mvbenchcomprehensivemultimodalvideo}, Next-QA\cite{xiao2021nextqanextphasequestionansweringexplaining}, and Video-MME\cite{fu2025videommefirstevercomprehensiveevaluation}. 

\subsubsection{Implementation Details}
For our ViTCoP framework, we configure the three-stage pruning process as follows: the first stage occurs at the output of the vision encoder, while the second and third stages are applied at the 2nd and 22nd layers of the LLM, respectively. For the VIC clustering algorithm, we set the distance threshold $d_c=8$ and the spatial threshold $\tau=0.6$. These hyperparameters were established based on preliminary experiments. They were kept fixed across all benchmarks without any dataset-specific fine-tuning to validate the robustness and strong generalization capabilities of our method. 
To ensure a fair comparison, all baseline methods adhere to their original experimental settings. All experiments were conducted on NVIDIA V100s GPUs, and all benchmarks were run using the \texttt{lmms-eval} package \cite{zhang2024lmmsevalrealitycheckevaluation}.

\subsection{Image-Language Understanding Tasks}

\begin{table}[ht]
\centering
\small
\renewcommand{\arraystretch}{0.8}
\setlength{\tabcolsep}{0.8pt} 

\begin{tabular}{lcccc c}
\toprule
\textbf{Method} & \textbf{EgoSch} & \textbf{MVB} & \textbf{Next-QA} & \textbf{V-MME} & \textbf{Avg (\%)} \\
\midrule
\textbf{Vanilla} & \perforiginal{0.414}{100.0\%} & \perforiginal{44.95}{100.0\%} & \perforiginal{26.64}{100.0\%} & \perforiginal{32.41}{100.0\%} & \textbf{100.0\%} \\
\midrule
\multicolumn{6}{c}{\textbf{Retain 128 Tokens {(↓ 88.9\%)}}} \\
\midrule
FastV & \perf{0.345}{83.2\%} & \perf{40.78}{90.7\%} & \perf{23.99}{90.1\%} & \perf{29.26}{90.3\%} & \peravg{88.6\%} \\
PyramidDrop & \perf{0.357}{86.3\%} & \perf{38.80}{86.3\%} & \perf{21.52}{80.8\%} & \perf{29.81}{92.0\%} & \peravg{86.4\%} \\
SparseVLM & \perfbest{0.406}{98.0\%} & \perfsecond{43.13}{96.0\%} & \perfsecond{24.77}{93.0\%} & \perfsecond{30.30}{93.5\%} & \peravg{95.1\%} \\
VisionZip & \perf{0.370}{89.3\%} & \perf{40.80}{90.8\%} & \perf{23.36}{87.7\%} & \perf{30.26}{93.4\%} & \peravg{90.3\%} \\
\cmidrule(lr){2-6}
\rowcolor{ourscell}
\textbf{ViTCoP (Ours)} & \perfsecond{0.405}{97.7\%} & \perfbest{43.30}{96.3\%} & \perfbest{25.60}{96.1\%} & \perfbest{32.67}{100.8\%} & \peravgbest{97.7\%} \\
\bottomrule
\end{tabular}
\caption{Performance on 4 video benchmarks from
LLaVA-NeXT-Video-7B.}
\label{tab:pruning_comparison_llava_video}
\end{table}

In this section, we systematically evaluate the performance and robustness of ViTCoP on two mainstream large vision-language models. We first conduct comprehensive tests on the LLaVA-1.5-7B model across 11 mainstream benchmark datasets. Subsequently, we further validate the scalability of ViTCoP under extreme compression scenarios on the higher-resolution LLaVA-NeXT-7B model.

\subsubsection{Performance on LLaVA-1.5-7B}

We evaluate the performance of ViTCoP under three pruning intensities: retaining 192 (66.7\% pruning), 128 (77.8\% pruning), and 64 (88.9\% pruning) tokens from the original 576 visual tokens. 
As shown in Table~\ref{tab:pruning_comparison_llava_1_5}, ViTCoP achieves the best average performance across all compression settings, significantly outperforming existing methods. 
For instance, at a moderate pruning rate (192 tokens), ViTCoP improves upon the next-best method, VisionZip, by 0.9\%. 
At an aggressive pruning of 64 tokens, ViTCoP still maintains 94.7\% performance, surpassing VisionZip and SparseVLM by 4.3\% and 14\%, respectively. 
It is worth noting that on some datasets, ViTCoP even exceeds the performance of the original model, reaching 100.3\% on MMBench and 100.4\% on OK-VQA. This suggests that our method not only effectively removes redundancy but can also mitigate the impact of interfering information on the model.

\subsubsection{Performance on LLaVA-NeXT-7B}

To verify the generalization capability of ViTCoP on high-resolution images, we conducted further experiments on the LLaVA-NeXT-7B model, which uses 2880 visual tokens, making more extreme pruning settings possible. We focused on evaluating two pruning rates: 88.9\% and 94.4\%. 
As shown in Table~\ref{tab:pruning_comparison_llava_next}, ViTCoP retains 95.1\% of the average performance at an 88.9\% pruning rate, significantly outperforming VisionZip's 91.4\%. 
At the more aggressive 94.4\% pruning rate, ViTCoP still achieves a 90.8\% retention rate, far exceeding VisionZip's 83.3\%. 
Notably, other methods such as FastV, PyramidDrop, and SparseVLM failed to run under this compression intensity and were therefore not included in the comparison. These results further validate the stability and strong generalization capability of ViTCoP under extreme compression conditions.

\subsection{Video-Language Understanding Tasks}

\begin{table}[ht]
\centering
\small
\renewcommand{\arraystretch}{0.8}
\setlength{\tabcolsep}{2pt} 
\begin{tabular}{l ccccc}
\toprule
\textbf{Ablation} & \textbf{TFLOPs} & \textbf{COCO} & \textbf{GQA} & \textbf{MMB} & \textbf{POPE} \\
\midrule
\rowcolor{ourscell}
\textbf{ViTCoP (Ours)} & 0.82 & 1.032 & \bestperf{0.574} & \bestperf{63.06} & \bestperf{0.807} \\
\midrule
w/o K-norm Guidance & 0.82 & 1.011 & 0.556 & 62.54 & 0.760 \\
w/o Attention Guidance & 0.82 & 1.015 & 0.563 & 62.63 & 0.771 \\
w/o Stage I Pruning & 0.91 & 0.086 & 0.389 & 20.27 & 0.283 \\
w/o Stage III Pruning & 0.81 & \bestperf{1.046} & 0.571 & 62.46 & 0.784 \\
\bottomrule
\end{tabular}
\caption{Ablation study. "w/o K-norm Guidance" uses only attention; "w/o Attention Guidance" uses only K-vector L2-norms. TFLOPs is avg. cost on COCO.}
\label{tab:component_ablation}
\end{table}

This section further evaluates the generalization and robustness of ViTCoP on dynamic temporal data. We extend the evaluation from static images to the video domain, conducting experiments with the LLaVA-NeXT-Video-7B model on four representative video question-answering datasets: EgoSchema, MVBench, Next-QA, and Video-MME. For these tasks, we uniformly apply an aggressive pruning rate of 88.9\%.

As shown in Table~\ref{tab:pruning_comparison_llava_video}, ViTCoP retains 96.3\% and 96.1\% of the performance on MVBench and Next-QA, respectively. In terms of average performance, ViTCoP (97.7\%) significantly outperforms SparseVLM (95.1\%) and achieves the best results on three of the four benchmarks.
ViTCoP's performance on Video-MME is particularly outstanding, reaching 100.8\% and even surpassing the original, unpruned model. 
Overall, our method achieves an average performance retention rate of 97.7\% across the four datasets, fully demonstrating ViTCoP's excellent generalization capabilities in video-language large models. 
These results indicate that ViTCoP not only excels in static image understanding but also maintains exceptional performance in complex video-language tasks, establishing it as a general and robust token pruning framework.

\subsection{Ablation Study}
\label{sec:ablation}

To evaluate ViTCoP's key components, we conduct an ablation study on the COCO, GQA, MMBench, and POPE datasets. We assess the multistage pruning and saliency metrics by creating four variants: \textit{w/o K-norm Guidance}, using only attention scores; \textit{w/o Attention Guidance}, using only the K-vector L2-norm; \textit{w/o Stage I Pruning}, removing the initial coarse-grained pruning; and \textit{w/o Stage III Pruning}, removing the final fine-grained pruning. Stage II is not ablated as it is integral to the pruning pipeline.

The ablation results in Table~\ref{tab:component_ablation} demonstrate that the full ViTCoP method significantly outperforms all variants. In particular, when using only the K-vector L2-norm as the saliency metric (\textit{w/o Attention Guidance}), performance does not degrade compared to \textit{w/o K-norm Guidance}, but it even slightly improves on some tasks. This highlights the K-vector L2-norm as an effective and robust proxy for token importance, with strong generalization and compatibility with modern acceleration techniques like FlashAttention. Additionally, the absence of visual guidance from VIC in the second stage, which relies only on the L2-norm for text-guided pruning, results in redundancy among retained key tokens, and thus degrades performance compared to the full ViTCoP.

\begin{table}[ht]
\centering
\small
\renewcommand{\arraystretch}{0.8}
\setlength{\tabcolsep}{0.5pt} 

\begin{tabular}{l ccccc}
\toprule
& \textbf{POPE} & \textbf{TFLOPs} & \textbf{GPU Mem} & \textbf{Prefill} & \textbf{Time/Tok} \\
\midrule
\textbf{LLaVA-NeXT} &
\perforiginal{0.863}{100\%} &
\perforiginal{31.55}{100\%} &
\perforiginal{30.80}{100\%} &
\perforiginal{914}{100\%} &
\perforiginal{62.67}{100\%} \\
\midrule
VisionZip &
\perfsecond{0.661}{76.6\%} &
\perfsecond{1.79}{5.7\%} &
\perfbest{27.12}{88.1\%} &
\perfbest{126}{13.8\%} &
\perfbest{53.39}{85.2\%} \\
\rowcolor{ourscell}
\textbf{ViTCoP (Ours)} &
\perfbest{0.755}{87.5\%} &
\perfbest{1.69}{5.4\%} &
\perfsecond{27.13}{88.1\%} &
\perfsecond{139}{15.2\%} &
\perfsecond{53.53}{85.4\%} \\
\bottomrule
\end{tabular}
\caption{Efficiency analysis of ViTCoP on LLaVA-NeXT-13B. Units: TFLOPs for computation, GB for GPU Memory, and ms for latency (Prefill and Time/Token).}
\label{tab:efficiency_comparison}
\end{table}

However, removing the first stage pruning (\textit{w/o Stage I Pruning}) resulted in a catastrophic performance decline. This outcome demonstrates that the initial removal of irrelevant tokens--such as redundant backgrounds, low-information regions, or repetitive textures--is crucial for alleviating the burden on subsequent pruning stages. Without this stage, the following stages struggle to discern redundancy, thereby retaining excessive noisy tokens that severely interfere with the model's representation capabilities.

Interestingly, removing the third stage pruning (\textit{w/o Stage III Pruning}) led to a slight improvement in the COCO dataset. This may be because the image-text matching task in COCO is highly sensitive to the aggregation of fine-grained visual semantics, and the further pruning in the third stage might inadvertently remove some detailed information, affecting the final performance.

In summary, the three stages of our method form a complementary and synergistic relationship. Removing any single stage leads to a performance drop or even significant degradation, highlighting the critical role of ViTCoP's multistage, progressive pruning strategy in achieving both effectiveness and robustness.

\subsection{Efficiency Analysis}
\label{efficiency}

ViTCoP achieves significant inference acceleration and computational savings by substantially reducing the number of visual tokens processed by the LLM. 
On the POPE dataset, we conduct a comparison based on LLaVA-NeXT-13B \cite{liu2024llavanext} against the vanilla model and VisionZip. 
As shown in Table~\ref{tab:efficiency_comparison}, ViTCoP reduces TFLOPs by over 94\%, decreases prefill latency by 85\%, and significantly shortens the generation time per token. 
Despite having efficiency comparable to VisionZip, ViTCoP demonstrates about 10\% higher performance retention, showcasing a superior trade-off between efficiency and performance.

\section{Conclusion}
\label{sec:conclusion}

In this paper, we introduce ViTCoP, a visual-textual semantic collaborative pruning framework designed to ensure that retained visual tokens are both crucial and informationally diverse. Extensive experiments on image and video understanding tasks demonstrate its effectiveness. ViTCoP maintains nearly 95\% of baseline performance at a high compression rate of 88.9\% and achieves performance retention of up to 97.7\% on video tasks, comprehensively outperforming existing state-of-the-art methods. As a tuning-free framework, ViTCoP reduces the TFLOPs of the model by more than 94\% while significantly reducing inference latency and GPU memory consumption. This offers a superior solution for the efficient deployment of Large Vision-Language Models in resource-constrained environments.

\bibliography{aaai2026}

\section*{Appendix A: Related Work}
\label{sec:related_work}

\subsection{Large Vision-Language Models}
\label{subsec:lvlm}

Large Vision-Language Models (LVLMs) have demonstrated remarkable success on a variety of multimodal tasks, such as image captioning and visual question answering \cite{huang2024ffaamultimodallargelanguage,lai2024lisareasoningsegmentationlarge,li2024minigeminiminingpotentialmultimodality,zhang2024prompt,zhou2023regionblip}. By integrating a vision encoder, such as a Vision Transformer (ViT), with a Large Language Model (LLM), models like LLaVA \cite{liu2023visualinstructiontuning}, Qwen-VL \cite{bai2023qwenvlversatilevisionlanguagemodel}, and InternVL \cite{chen2024internvlscalingvisionfoundation} have achieved state-of-the-art performance through effective vision-language alignment modules. 
However, a recent trend in LVLM research is the shift towards processing higher-resolution inputs, including both images and videos. For instance, LLaVA-NeXT \cite{liu2024llavanext} generates nearly 3,000 visual tokens from a single $672 \times 672$ pixel image. Similarly, models like LLaVA-Video \cite{zhang2024llavanextvideo}, which process sequences of video frames, can result in input lengths scaling to tens of thousands of tokens. This dramatic increase in sequence length leads to a substantial rise in inference latency and computational memory overhead. Consequently, developing efficient computational paradigms for LVLMs has become a critical and pressing research problem.

\subsection{Visual Token Pruning}
\label{subsec:pruning}

To enhance inference efficiency, several methods have been proposed to prune redundant Key-Value (KV) caches in language models, such as H2O \cite{zhang2023ho} and StreamingLLM \cite{xiao2024efficient}. Recently, similar strategies have been extended to manage the significantly larger visual token sequences in LVLMs \cite{chen2024imageworth12tokens,he2024zipvlefficientlargevisionlanguage,pmlr-v202-shi23e,xing2025pyramiddropacceleratinglargevisionlanguage,zhang2025sparsevlmvisualtokensparsification,yang2024visionziplongerbetternecessary}.
Existing approaches to visual token pruning can be broadly categorized into two main types. The first category performs pruning within the vision encoder, utilizing attention-based or clustering-driven techniques \cite{bolya2023token, yang2024visionziplongerbetternecessary}. While computationally efficient, these methods often lack semantic guidance from the language model, leading to the risk of erroneously discarding critical visual information. The second category implements pruning inside the LLM based on cross-attention scores from textual queries \cite{chen2024imageworth12tokens,zhang2025sparsevlmvisualtokensparsification}. This allows the model to focus on semantically relevant tokens, but it frequently results in a subset with high redundancy, as many tokens may share similar semantic attributes.
To address these limitations, this paper introduces ViTCoP, a Visual-Text Semantic Co-Pruning Framework. ViTCoP is designed to select a token subset that is both representative and diverse, thereby improving inference efficiency while simultaneously enhancing model performance.

\section*{Appendix B: Evaluation Benchmarks and Metrics}

This section details the datasets and evaluation protocols used in our experiments. To comprehensively assess model performance, we selected 11 image-language and 4 video-language benchmarks covering a diverse range of tasks, including Image Captioning, Visual Reasoning, and various formats of Visual Question Answering (VQA).

We employ standard metrics for each task. For image captioning on COCO-2017, Flickr30k, and Nocaps, we report the CIDEr score \cite{7299087}. For VQA tasks, metrics vary by format: Accuracy is used for multiple-choice benchmarks on both images (MMBench, QBench) and videos (EgoSchema, MVBench). Exact Match is used for reasoning and closed-ended QA on ScienceQA, GQA, OK-VQA, and VQA-v2. To evaluate object hallucination on POPE, we use the F1 Score. Open-ended QA on Next-QA is evaluated with the Wu-Palmer Similarity (WUPS) score \cite{wu1994verbsemanticslexicalselection}. Finally, the Perception Score \cite{fu2024mmecomprehensiveevaluationbenchmark} is used for both MME and Video-MME.

Our evaluation uses standard data splits, such as Validation or Test. In the accompanying table, Default indicates the use of a standard test split or that only a single split is available. For all performance metrics, a higher value indicates better performance. Table~\ref{tab:datasets_revised} provides a detailed summary of each dataset, task, and metric.

\begin{table*}[t!]
\centering
\small 
\caption{An overview of the datasets used for evaluation. We list the dataset, task type, evaluation metric, and the specific data split and subset used.}
\label{tab:datasets_revised}
\begin{tabular*}{\textwidth}{@{\extracolsep{\fill}}lllll@{}}
\toprule
\textbf{Dataset} & \textbf{Task} & \textbf{Metric} & \textbf{Split} & \textbf{Subset} \\ \midrule
\multicolumn{5}{c}{\textbf{Image-Language Datasets}} \\ \midrule
COCO-2017~\cite{lin2015microsoft} & Image Captioning & CIDEr & Validation & Full \\
Flickr30k~\cite{young-etal-2014-image} & Image Captioning & CIDEr & Test & Full \\
GQA~\cite{hudson2019gqa} & Closed-Ended VQA & Exact Match & Default & Full \\
MMBench~\cite{liu2024mmbenchmultimodalmodelallaround} & Multiple-Choice VQA & Accuracy & Validation & English \\
MME~\cite{fu2024mmecomprehensiveevaluationbenchmark} & Closed-Ended VQA & Perception Score & Default & Full \\
Nocaps~\cite{Agrawal_2019} & Image Captioning & CIDEr & Validation & Full \\
OK-VQA~\cite{marino2019okvqavisualquestionanswering} & Visual Reasoning & Exact Match & Validation & Full \\
POPE~\cite{li2023evaluating} & Closed-Ended VQA & F1 Score & Default & Full \\
QBench~\cite{wu2024qbenchbenchmarkgeneralpurposefoundation} & Multiple-Choice VQA & Accuracy & Test & Full \\
ScienceQA~\cite{lu2022learn} & Visual reasoning & Exact Match & Test & Vision only \\
VQA-v2~\cite{8100153} & Closed-Ended VQA & Exact Match & Validation & Lite \\ \midrule
\multicolumn{5}{c}{\textbf{Video-Language Datasets}} \\ \midrule
EgoSchema~\cite{mangalam2023egoschemadiagnosticbenchmarklongform} & Multiple-Choice VQA & Accuracy & Test & MC/Subset \\
MVBench~\cite{li2024mvbenchcomprehensivemultimodalvideo} & Multiple-Choice VQA & Accuracy & Default & Full \\
Next-QA~\cite{xiao2021nextqanextphasequestionansweringexplaining} & Open-Ended VQA & WUPS & Test & OE \\
Video-MME~\cite{fu2025videommefirstevercomprehensiveevaluation} & Closed-Ended VQA & Perception Score & Default & Full \\ \bottomrule
\end{tabular*}
\end{table*}

\section*{Appendix C: Implementation Details of ViTCoP}

In all image-language and video-language tasks, we configure the first stage of pruning in the ViTCoP framework to occur at the penultimate layer of the visual encoder. This means pruning is performed just before the visual tokens are fed into the projection layer, a setup that follows the original configuration of LVLM models. 

For the second and third stages of pruning, we conduct them within the LLM at layers 2 and 22, respectively. The selection of these specific shallow and deep layers was based on a dedicated study investigating the impact of applying pruning at different depths within the LLM, with the results presented in Table~\ref{tab:layer_ablation}. This study on pruning layer selection reveals that model performance is highly sensitive to these positions. For instance, selecting layer 1 for shallow pruning results in a performance drop because the LLM has not yet sufficiently converged the visual token information. This leads to a lack of textual semantic guidance for the vision-text co-pruning, preventing the model from capturing key diverse visual information and aggregating global context effectively. Conversely, applying deep pruning too early, such as at layer 17, causes critical visual information to be discarded before it can be fully absorbed by the model, leading to a severe degradation in performance. Therefore, our default configuration places the shallow, vision-text co-guided pruning at layer 2 and the deep, text-guided pruning at layer 22. This setup achieves the most effective balance between high performance and computational efficiency.

For the VIC clustering algorithm, we set the distance threshold $d_c$ and the spatial threshold $\tau$. To ensure the generalizability of the algorithm, we determined these fixed parameters by performing a grid search exclusively on the COCO-2017 dataset \cite{lin2015microsoft}, selecting $d_c=8$ and $\tau=0.6$. The token retention ratios for each stage of ViTCoP at different overall pruning rates on the LLaVA-1.5-7B model \cite{liu2023visualinstructiontuning} are detailed in Table~\ref{tab:retention_ratios}.

\begin{table*}[ht]
\centering
\renewcommand{\arraystretch}{1.25}
\caption{Ablation study on the selection of shallow ($l_s$) and deep ($l_d$) pruning layers within the LLM. The best performance for each metric is highlighted in bold. Our chosen configuration provides the best overall trade-off.}
\label{tab:layer_ablation}
\begin{tabular*}{\textwidth}{@{\extracolsep{\fill}}l cccccccc}
\toprule
\textbf{Pruning Layers} & \textbf{TFLOPs} & \textbf{COCO} & \textbf{GQA} & \textbf{MMBench} & \textbf{MME} & \textbf{OK-VQA} & \textbf{POPE} & \textbf{VQA-v2} \\
\midrule
\rowcolor{ourscell}
\textbf{$l_s=2,l_d=22$ (Ours)} & 0.82 & 1.0315 & \bestperf{0.5741} & \bestperf{63.06} & \bestperf{1744} & \bestperf{0.5084} & \bestperf{0.8069} & 0.6632 \\
\midrule
$l_s=1,l_d=22$ & 0.82 & 1.0143 & 0.5655 & 62.11 & 1713 & 0.5032 & 0.8032 & 0.6430 \\
$l_s=3,l_d=22$ & 0.82 & 1.0301 & 0.5707 & 62.97 & 1728 & 0.5056 & 0.8053 & \bestperf{0.6692} \\
$l_s=2,l_d=17$ & 0.82 & 0.8745 & 0.5537 & 62.89 & 1734 & 0.4682 & 0.7980 & 0.5966 \\
$l_s=2,l_d=27$ & 0.82 & \bestperf{1.0347} & 0.5717 & 62.80 & 1740 & 0.5009 & 0.7903 & 0.6644 \\
\bottomrule
\end{tabular*}%
\end{table*}

\begin{table*}[ht]
\centering
\caption{Token retention ratios at each stage for different overall pruning rates on LLaVA-1.5-7B.}
\label{tab:retention_ratios}
\renewcommand{\arraystretch}{1.2}
\setlength{\tabcolsep}{5.5pt} 
\begin{tabular}{@{}c c c c c c c c c c c c@{}}
\toprule
\multicolumn{3}{c}{\textbf{Pruning 66.7\%}} & \multicolumn{3}{c}{\textbf{Pruning 77.8\%}} & \multicolumn{3}{c}{\textbf{Pruning 88.9\%}} & \multicolumn{3}{c}{\textbf{Pruning 94.4\%}} \\
\cmidrule(r){1-3} \cmidrule(r){4-6} \cmidrule(r){7-9} \cmidrule(l){10-12}
Stage I & Stage II & Stage III & Stage I & Stage II & Stage III & Stage I & Stage II & Stage III & Stage I & Stage II & Stage III \\
\midrule
0.5000 & 0.4394 & 0.0879 & 0.4000 & 0.2869 & 0.0574 & 0.3000 & 0.1343 & 0.0269 & 0.2500 & 0.0581 & 0.0116 \\
\bottomrule
\end{tabular}
\end{table*}

\section{Appendix D: Additional Algorithm Details}
\label{sec:appendix_algorithms}

This appendix provides detailed pseudocode for the key components of the ViTCoP framework. Algorithm~\ref{alg:vitcop_framework} outlines the complete three-stage pruning process. Algorithm~\ref{alg:vic} details the Visual Information Clustering (VIC) method used in Stage II for preserving semantic diversity. Algorithm~\ref{alg:collaborative_pruning} specifies the collaborative pruning and merging strategy, also from Stage II, which synergizes visual diversity with textual relevance.

\begin{algorithm}[ht]
\caption{ViTCoP: Visual-Textual Collaborative Pruning Framework}
\label{alg:vitcop_framework}
\begin{algorithmic}[1]
\State \textbf{Input:} Initial visual tokens $V_{in}$, Text query tokens $T_{in}$, Pruning schedule $\mathcal{S}$, Pruning ratios $\{\pi_1, \pi_2, \pi_3\}$.
\State \textbf{Output:} Final visual tokens $V_{out}$.
\State $m \gets |V_{in}|$ \Comment{Store original number of visual tokens}

\Statex \Comment{Stage I: Visual Saliency-Guided Pruning in Vision Encoder}
\State At the specified layer in $\mathcal{S}$ for Stage I:
\State Calculate saliency scores $S$ for each token in $V_{in}$.
\State $k_1 \gets \lfloor m \cdot \pi_1 \rfloor$ \Comment{Target count relative to original m}
\State $V_{stage1} \gets \text{TopK}(V_{in}, \text{scores=}S, \text{k=}k_1)$.
\State Let $V_{current} \gets V_{stage1}$.

\Statex \Comment{Project tokens and feed into LLM}
\State $V_{current} \gets \text{ProjectionLayer}(V_{current})$.
\State The sequence for the LLM is $[V_{current}, T_{in}]$.

\Statex \Comment{Stages II \& III within LLM Layers}
\For{each LLM layer $l=1, ..., L$}
    \If{$l$ is a specified layer in $\mathcal{S}$ for Stage II}
        \Statex \Comment{Stage II: Visual-Textual Collaborative Pruning}
        \State Let $F$ be the feature vectors and $P$ be the position vectors of tokens in $V_{current}$.
        \State $L \gets \text{VIC}(F, P)$ \Comment{Run Algorithm 2}
        \State Calculate K-vector L2 norms $K_{norms}$ for tokens in $V_{current}$.
        \State $B \gets \lfloor m \cdot \pi_2 \rfloor$ \Comment{Target count relative to original m}
        \State $\begin{aligned}[t]
    V_{current} \gets \text{CollaborativePruning}( & V_{current}, L, \\
                                                  & K_{norms}, B)
\end{aligned}$ \Comment{Run Algorithm 3}
    \ElsIf{$l$ is a specified layer in $\mathcal{S}$ for Stage III}
        \Statex \Comment{Stage III: Textual Saliency-Guided Pruning}
        \State Calculate K-vector L2 norms $K_{norms}$ for tokens in $V_{current}$.
        \State $k_3 \gets \lfloor m \cdot \pi_3 \rfloor$ \Comment{Target count relative to original m}
        \State $V_{current} \gets \text{TopK}(V_{current}, \text{scores=}-K_{norms}, \text{k=}k_3)$ \Comment{Smaller norm is better}
    \EndIf
    \State Process the token sequence through layer $l$.
\EndFor
\State $V_{out} \gets V_{current}$.
\State \textbf{return} $V_{out}$
\end{algorithmic}
\end{algorithm}

\begin{algorithm}[ht]
\caption{Visual Information Clustering (VIC)}
\label{alg:vic}
\begin{algorithmic}[1]
\State \textbf{Input:} Feature set $F = \{f_1, ..., f_n\}$, Position set $P = \{p_1, ..., p_n\}$, cutoff distance $d_c$, spatial threshold $\tau$, center ratio $\alpha$.
\State \textbf{Output:} Cluster labels array $L$.
\Statex \Comment{Step 1: Compute pairwise distances and local densities}
\State $D_{feat} \gets \text{ComputePairwiseDistances}(F)$
\State $D_{spatial} \gets \text{ComputePairwiseDistances}(P)$
\For{each token $i \in \{1, ..., n\}$}
    \State $\rho_i \gets \sum_{j \neq i} \exp\left(-\left(D_{feat}[i,j]/d_c\right)^2\right)$
\EndFor
\Statex
\Statex \Comment{Step 2: Compute minimum distance to higher density neighbors}
\For{each token $i \in \{1, ..., n\}$}
    \State $\delta_i \gets \infty$
    \State $N_i \gets -1$ \Comment{Index of nearest higher-density neighbor}
    \For{each token $j \in \{1, ..., n\}$}
        \If{$\rho_j > \rho_i$ and $D_{spatial}[i,j] \leq \tau$}
            \If{$D_{feat}[i,j] < \delta_i$}
                \State $\delta_i \gets D_{feat}[i,j]$
                \State $N_i \gets j$
            \EndIf
        \EndIf
    \EndFor
\EndFor
\Statex
\Statex \Comment{Step 3: Identify cluster centers based on importance score $\gamma$}
\State $\gamma \gets \rho \odot \delta$ \Comment{Element-wise product}
\State $N_{centers} \gets \lceil n \cdot \alpha \rceil$
\State $C_{indices} \gets \text{indices of top } N_{centers} \text{ values in } \gamma$
\Statex
\Statex \Comment{Step 4: Assign cluster labels}
\State Initialize $L$ of size $n$ with $-1$.
\For{$k$ from $1$ to $N_{centers}$}
    \State $L[C_{indices}[k]] \gets k$ \Comment{Assign unique label to each center}
\EndFor
\State $S_{indices} \gets \text{indices of tokens sorted by } \rho \text{ in descending order}$
\For{each index $i$ in $S_{indices}$}
    \If{$L[i] = -1$} \Comment{If token is not a center}
        \State $L[i] \gets L[N_i]$ \Comment{Assign label of its nearest higher-density parent}
    \EndIf
\EndFor
\State \textbf{return} $L$
\end{algorithmic}
\end{algorithm}

\begin{algorithm}[ht]
\caption{Visual-Textual Collaborative Pruning and Merging}
\label{alg:collaborative_pruning}
\begin{algorithmic}[1]
\State \textbf{Input:} Visual tokens $V=\{v_1, ..., v_n\}$, Cluster labels $L=\{l_1, ..., l_n\}$, Key L2 norms $K_{norms}$, Target token budget $B$.
\State \textbf{Output:} Pruned and merged visual tokens $V_{out}$.
\State $V_{elites} \gets \emptyset$, $V_{merged} \gets \emptyset$
\Statex \Comment{Group tokens by cluster label}
\State Clusters $C \gets \text{GroupTokensByLabel}(V, L)$ \Comment{$C=\{C_1, ..., C_{N_c}\}$}
\State $N_c \gets |C|$
\State $B_{elites} \gets B - N_c$ \Comment{Budget for elite tokens}
\Statex
\Statex \Comment{Retain elite tokens and merge the rest for each cluster}
\For{each cluster $C_c$ in $C$}
    \Statex \Comment{Allocate retention quota for elites based on relative cluster size}
    \State $q_c \gets \left\lfloor \frac{|C_c|}{n} \cdot B_{elites} \right\rfloor$ 
    \State Sort tokens in $C_c$ by their $K_{norms}$ in ascending order.
    \State $E_c \gets \text{first } q_c \text{ tokens from sorted } C_c$.
    \State $V_{elites} \gets V_{elites} \cup E_c$
    \Statex
    \Statex \Comment{Merge remaining non-elite tokens in the cluster}
    \State $C_c^{\text{remaining}} \gets C_c \setminus E_c$
    \If{$C_c^{\text{remaining}} \neq \emptyset$}
        \State $v_{merged} \gets \frac{1}{|C_c^{\text{remaining}}|} \sum_{v_i \in C_c^{\text{remaining}}} v_i$ 
        \State $V_{merged} \gets V_{merged} \cup \{v_{merged}\}$
    \EndIf
\EndFor
\State $V_{out} \gets V_{elites} \cup V_{merged}$
\State \textbf{return} $V_{out}$
\end{algorithmic}
\end{algorithm}

\section{Appendix E: Theoretical Analysis of Computational Complexity}
\label{sec:appendix_complexity}

This appendix provides a theoretical analysis of the computational overhead introduced by ViTCoP and the corresponding reduction in inference FLOPs.

\subsection{Algorithm Complexity Analysis}
The computational overhead of ViTCoP primarily stems from the three pruning stages. Let $N_v$ be the number of visual tokens input to a given stage.

\paragraph{Stage I: Visual Saliency-Guided Pruning.}
This stage is executed once in the vision encoder. Its complexity is dominated by selecting the top-$k$ tokens from the initial set of $m$ visual tokens, making it approximately $O(m \log m)$.

\paragraph{Stage II: Visual-Textual Collaborative Pruning.}
This stage is applied at layer $l_s$ on $n_{s} = \pi_1 \cdot m$ tokens. The complexity is dominated by the VIC algorithm's pairwise distance calculations, resulting in a total complexity of $O((\pi_1 m)^2)$.

\paragraph{Stage III: Textual Saliency-Guided Pruning.}
Applied at the deep layer $l_d$ on $n_{d} = \pi_2 \cdot m$ tokens, this stage has a complexity of $O(n_d \log n_d)$.

\subsection{TFLOPs Reduction in LLM Inference}
The computational cost (FLOPs) of a transformer layer for a sequence of length $N$ and hidden dimension $d$ is approximately:
\begin{equation}
\label{eq:flops_layer}
\text{FLOPs}_{\text{layer}}(N) \approx \underbrace{4N^2d}_{\text{Attention}} + \underbrace{16Nd^2}_{\text{FFN}}
\end{equation}
ViTCoP reduces the number of visual tokens ($N_v$) progressively. Let $m$ be the original number of visual tokens and $N_t$ be the number of text tokens. The visual token count $N_v(l)$ at layer $l$ is:
\begin{equation}
\label{eq:nv_l}
N_v(l) = 
\begin{cases} 
\pi_1 \cdot m & \text{if } 1 \le l < l_s \\
\pi_2 \cdot m & \text{if } l_s \le l < l_d \\
\pi_3 \cdot m & \text{if } l_d \le l \le n 
\end{cases}
\end{equation}
where $n$ is the total number of LLM layers. The total FLOPs with ViTCoP are calculated as:
\begin{equation}
\label{eq:flops_vitcop}
\begin{split}
\text{FLOPs}_{\text{ViTCoP}} = & (l_s-1) \cdot \text{FLOPs}_{\text{layer}}(\pi_1 m + N_t) \\
& + (l_d - l_s) \cdot \text{FLOPs}_{\text{layer}}(\pi_2 m + N_t) \\
& + (n - l_d + 1) \cdot \text{FLOPs}_{\text{layer}}(\pi_3 m + N_t)
\end{split}
\end{equation}
Since $\pi_3 < \pi_2 < \pi_1 < 1$, the FLOPs reduction is substantial.

\subsection{Integrated Token Compression Ratio}
To holistically measure efficiency, we define an Integrated Token Compression Ratio ($CR_{\text{int}}$) that averages the token reduction over all $n$ layers of the LLM.
\begin{multline}
\label{eq:cr_int}
CR_{\text{int}} = 1 - \frac{1}{n \cdot m} \Big( (\pi_1 m)(l_s-1) \\
+ (\pi_2 m)(l_d - l_s) + (\pi_3 m)(n - l_d + 1) \Big)
\end{multline}
This metric accurately reflects the overall reduction in computational workload.

\section*{Appendix F: Visual Token Importance Across LLM Layers}

\begin{figure*}[ht]
\centering
\includegraphics[width=0.9\linewidth]{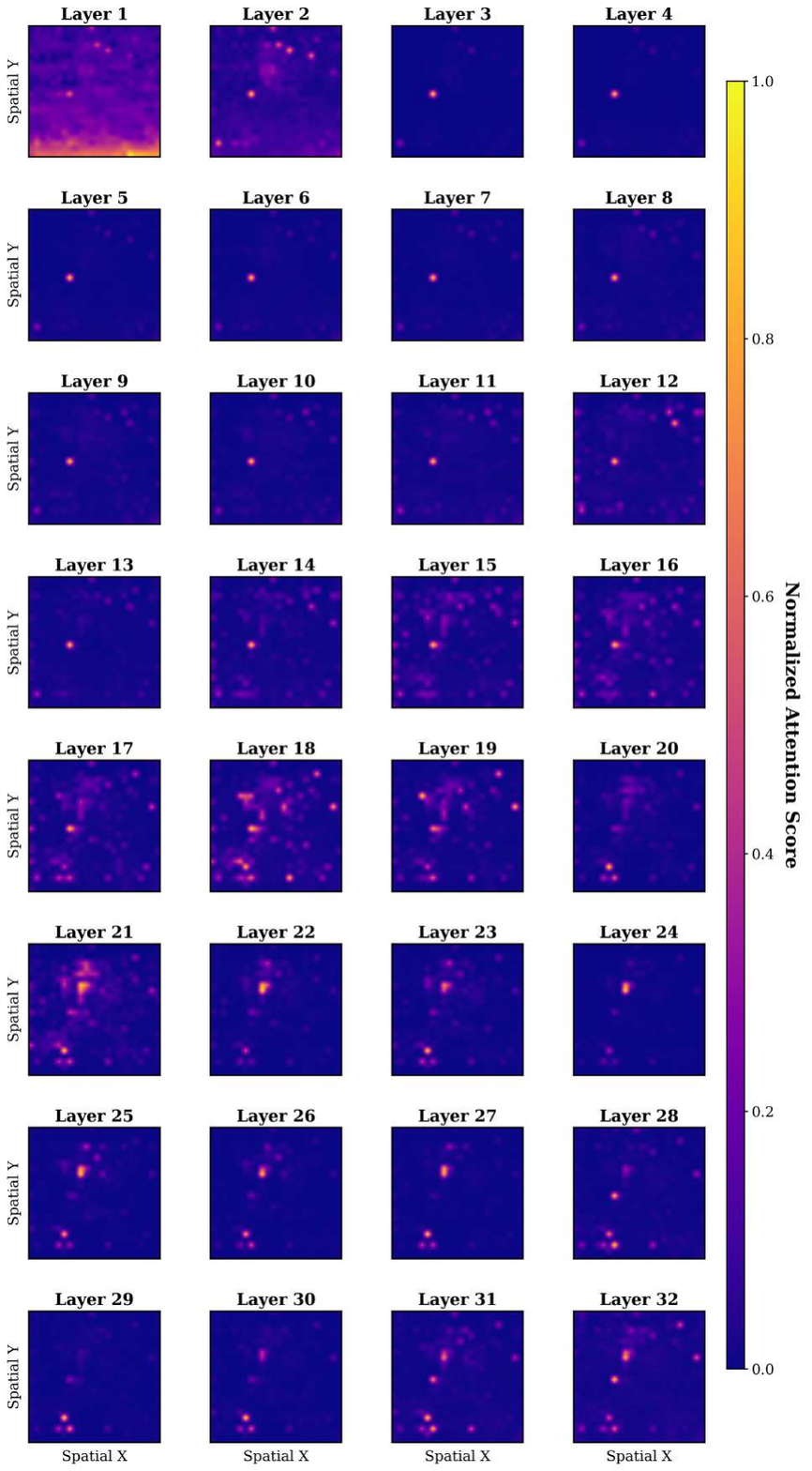}
\caption{Heatmaps of visual token attention distribution across all layers of the LLaVA-1.5-7B LLM on the COCO dataset. The attention patterns evolve from being diffuse and global in shallow layers (top rows) to sparse and highly focused in deep layers (bottom rows). This illustrates a functional transition from coarse-grained visual aggregation to fine-grained local detail absorption.}
\label{fig:appendix_attention_heatmap}
\end{figure*}

This appendix provides a detailed visualization of visual token attention scores across all 32 layers of the Large Language Model (LLM), supplementing the analysis of Key Insight 3 in the main text. Figure \ref{fig:appendix_attention_heatmap} illustrates the full evolutionary trajectory of attention paid to visual tokens as they are processed through the model.

The layer-by-layer heatmaps provide granular evidence for how the LLM dynamically shifts its focus.

This observed trajectory provides strong evidence that visual tokens undergo a process of progressive refinement within the LLM. The model first builds a comprehensive understanding of the global visual scene and then narrows its focus to absorb the most critical local details, demonstrating an efficient and dynamic allocation of computational resources.




\end{document}